\documentclass{article}

\pdfoutput=1
\usepackage{microtype}
\usepackage{graphicx}
\usepackage{svg}
\usepackage{subfig}
\usepackage{amsfonts}
\usepackage{paralist}
\usepackage{verbatim}
\usepackage{booktabs} 
\usepackage{amsmath}
\usepackage{gensymb}
\usepackage{multirow}

\usepackage{hyperref}

\newcommand{\norm}[1]{\left\lVert#1\right\rVert}
\newcommand{\Lagr}{\mathcal{L}}
\newcommand*{\pd}[3][]{\ensuremath{\frac{\partial^{#1} #2}{\partial #3}}}

\usepackage[accepted]{icml2020}

\icmltitlerunning{DisCont: Self-Supervised Visual Attribute Disentanglement using Context Vectors}

\begin{document}

\twocolumn[
\icmltitle{DisCont: Self-Supervised Visual Attribute Disentanglement \\ using Context Vectors}



\icmlsetsymbol{equal}{*}

\begin{icmlauthorlist}
\icmlauthor{Sarthak Bhagat}{equal,to}
\icmlauthor{Vishaal Udandarao}{equal,to}
\icmlauthor{Shagun Uppal}{equal,to}
\end{icmlauthorlist}

\icmlaffiliation{to}{IIIT Delhi, New Delhi, India}

\icmlcorrespondingauthor{Sarthak Bhagat}{sarthak16189@iiitd.ac.in}
\icmlcorrespondingauthor{Vishaal Udandarao}{vishaal16119@iiitd.ac.in}
\icmlcorrespondingauthor{Shagun Uppal}{shagun16088@iiitd.ac.in}

\icmlkeywords{Machine Learning, ICML}

\vskip 0.3in
]



\printAffiliationsAndNotice{\icmlEqualContribution \\} 

\begin{abstract}
Disentangling the underlying feature attributes within an image with no prior supervision is a challenging task. Models that can disentangle attributes well provide greater interpretability and control. In this paper, we propose a self-supervised framework \textit{DisCont} to disentangle multiple attributes by exploiting the structural inductive biases within images. Motivated by the recent surge in contrastive learning paradigms, our model bridges the gap between self-supervised contrastive learning algorithms and unsupervised disentanglement. We evaluate the efficacy of our approach, both qualitatively and quantitatively, on four benchmark datasets.
\end{abstract}

\section{Introduction}
Real world data like images are generated from several independent and interpretable underlying attributes \cite{deep_rep}. It has generally been assumed that successfully disentangling these attributes can lead to robust task-agnostic representations which can enhance efficiency and performance of deep models \cite{causal, bengio2013representation, peters2017elements}. However, recovering these independent factors in a completely unsupervised manner has posed to be a major challenge.

Recent approaches to unsupervised disentanglement have majorly used variants of variational autoencoders \cite{Higgins2017betaVAELB, Kim2018DisentanglingBF, Chen2018IsolatingSO, Kim2019RelevanceFV} and generative adversarial networks \cite{infogan, Hu2018DisentanglingFO, Shukla2019ProductOO}. Further, such disentangled representations have been utilized for a diverse range of applications including domain adaptation \cite{Cao2018DiDADS, Vu2019DomainAM, Yang2019UnsupervisedDA}, video frame prediction \cite{Denton2017UnsupervisedLO, Villegas2017DecomposingMA, Hsieh2018LearningTD, Bhagat2020DisentanglingRU}, recommendation systems \cite{Ma2019LearningDR} and multi-task learning \cite{Meng2019RepresentationDF}.

Contrary to these approaches, \cite{loca} introduced an `impossibility result' which showed that unsupervised disentanglement is impossible without explicit inductive biases on the models and data used. They empirically and theoretically proved that without leveraging the implicit structure induced by these inductive biases within various datasets, disentangled representations cannot be learnt in an unsupervised fashion.

Inspired by this result, we explore methods to exploit the spatial and structural inductive biases prevalent in most visual datasets \cite{inductive_bias_1, inductive_bias_2}. Recent literature on visual self-supervised representation learning  \cite{PIRL, CMC, moco, theoretical_contrastive, SimCLR} has shown that using methodically grounded data augmentation techniques using contrastive paradigms \cite{NCE, CPC, CPC2} is a promising direction to leverage such inductive biases present in images. The success of these contrastive learning approaches in diverse tasks like reinforcement learning \cite{structured_world_models, curl}, multi-modal representation learning \cite{multi-modal-contrast, cobra} and information retrieval \cite{deepchannel, contravis} further motivates us to apply them to the problem of unsupervised disentangled representation learning.


In this work, we present an intuitive self-supervised framework \textit{DisCont} to disentangle multiple feature attributes from images by utilising meaningful data augmentation recipes. We hypothesize that applying various stochastic transformations to an image can be used to recover the underlying feature attributes. Consider the example of data possessing two underlying attributes, \textit{i.e}, color and position. To this image, if we apply a color transformation (\textit{eg.} color jittering, gray-scale transform), only the underlying color attribute should change but the position attribute should be preserved. Similarly, on applying a translation and/or rotation to the image, the position attribute should vary keeping the color attribute intact.

It is known that there are several intrinsic variations present within different independent attributes \cite{describing_attributes, zhang2019multi}. To aptly capture these variations, we introduce `Attribute Context Vectors' (refer Section \ref{attribute_cv}).
We posit that by constructing attribute-specific context vectors that learn to capture the entire variability within that attribute, we can learn richer and more robust representations. 

Our major contributions in this work can be summarised as follows:
\begin{compactitem}
    \setlength{\itemsep}{1pt}
    \setlength{\parskip}{0pt}
    \setlength{\parsep}{0pt}
    \item We propose a self-supervised method \textit{DisCont} to simultaneously disentangle multiple underlying visual attributes by effectively introducing inductive biases in images via data augmentations.
    \item We highlight the utility of leveraging composite stochastic transformations for learning richer disentangled representations.
    \item We present the idea of `Attribute Context Vectors' to capture and utilize intra-attribute variations in an extensive manner. 
    \item We impose an attribute clustering objective that is commonly used in distance metric learning literature, and show that it further promotes attribute disentanglement. 
\end{compactitem}

The rest of the paper is organized as follows: Section \ref{methodology} presents our proposed self-supervised attribute disentanglement framework, Section \ref{experiments} provides empirical verification for our hypotheses using qualitative and quantitative evaluations, and Section \ref{conclusion} concludes the paper and provides directions for future research.

\section{Methodology}
\label{methodology}
In this section, we start off by introducing the notations we follow, move on to describing the network architecture and the loss functions employed, and finally, illustrate the training procedure and optimization strategy adopted.

\subsection{Preliminaries}
Assume we have a dataset $X = \{x_1, x_2, ..., x_N\}$ containing $N$ images, where $x_i \in \mathbb{R}^{C \times H \times W}$, consisting of $K$ labeled attributes $y = \{y_{1}, y_{2} ... y_{K}\}$. These images can be thought of as being generated by a small set of explicable feature attributes. For example, consider the CelebA dataset \cite{celeba} containing face images. A few of the underlying attributes are hair color, eyeglasses, bangs, moustache, \textit{etc}. 

\begin{figure}[!t]
    \centering
    \includegraphics[scale=0.48]{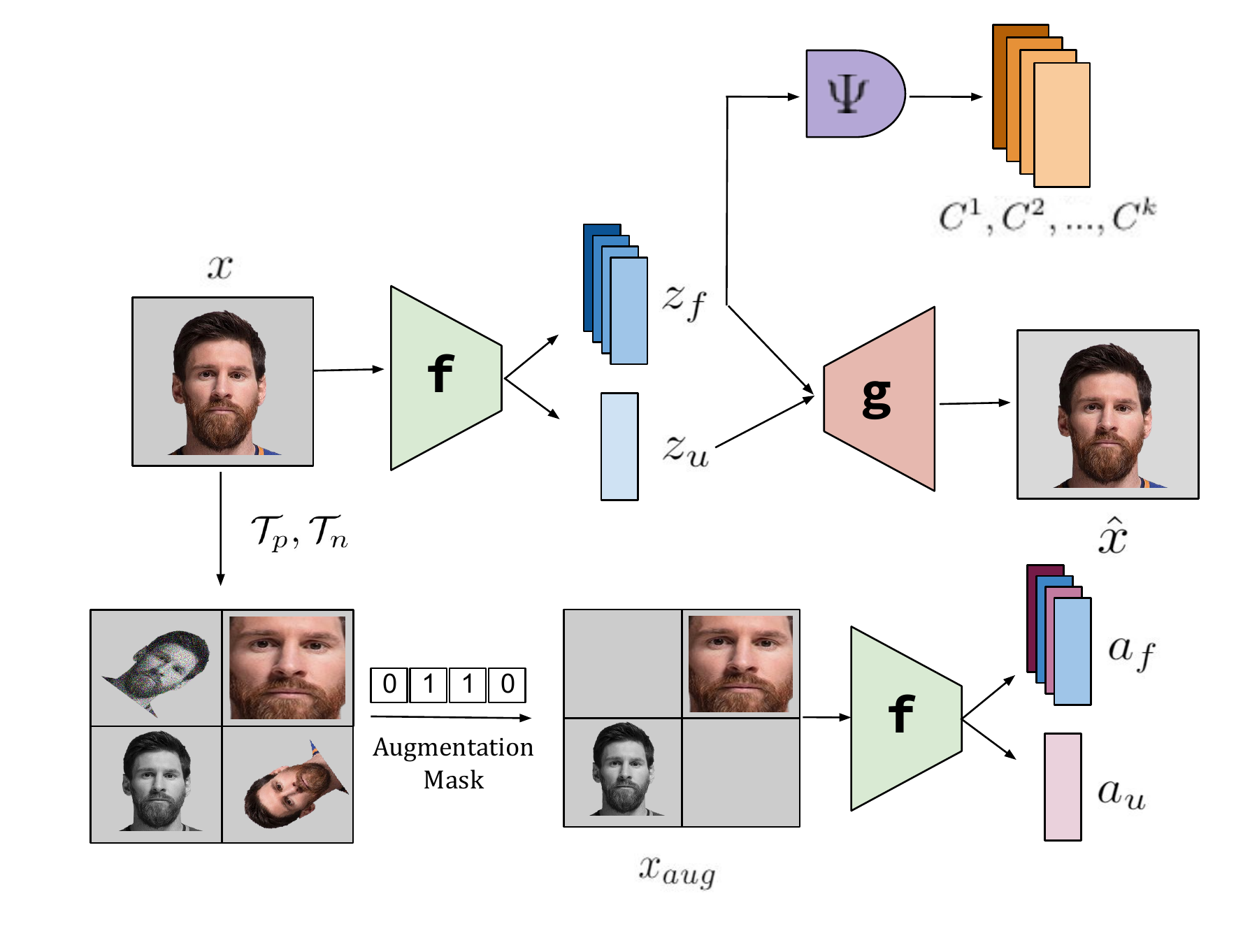}
    \vspace{-0.55cm}
    \caption{Overview of our architecture \textit{DisCont}. Given a batch of images $x$, we generate an augmented batch $x_{aug}$ by sampling a set of stochastic transformations. We then encode $x$ and $x_{aug}$ to extract their latent representations $z_f$, $z_u$ and $a_f$, $a_u$ respectively. $z_f$ is then used to construct `Attribute Context Vectors' $C^{1}, C^{2}, ..., C^{k}$ corresponding to each feature attribute. The context vectors and the latent representations are then used to optimize our disentanglement objective.}
    \label{fig:model}
\end{figure}

From \cite{Do2020TheoryAE}, we define a latent representation chunk $z_{i}$ as `fully disentangled' \textit{w.r.t} a ground truth factor $y_{k}$ if $z_{i}$ is fully separable from $z_{\ne i}$ and $z_{i}$ is fully interpretable \textit{w.r.t} $y_{k}$. Therefore, we can say for such a representation, the following conditions hold:
\vspace{-0.2cm}
\begin{equation}
    \mathbb{I}(z_{i}, z_{\ne i}) = 0 
\end{equation}
and 
\begin{equation}
\mathbb{I}(z_{i}, y_{k}) = \mathbb{H}(z_{i}, y_{k})    
\end{equation}

where, $\mathbb{I}(\cdot, \cdot)$ denotes the mutual information between two latent chunks while $\mathbb{H}(\cdot, \cdot)$ denotes the entropy of the latent chunk \textit{w.r.t} attribute.
To recover these feature attributes in a self-supervised manner while ensuring attribute disentanglement, we propose an encoder-decoder network (refer Fig \ref{fig:model}) that makes use of contrastive learning paradigms.

\subsection{Model Description}
To enforce the learning of rich and disentangled attributes, we propose to view the underlying latent space as two disjoint subspaces. 
\begin{compactitem}
    \setlength{\itemsep}{1pt}
    \setlength{\parskip}{0pt}
    \setlength{\parsep}{0pt}
    \item $Z_f \subset \mathbb{R}^{d \times k}$: denotes the feature attribute space containing the disentangled and interpretable attributes, where $d$ and $k$ denote the dimensionality of the space and the number of feature attributes respectively.
    \item $Z_u \subset \mathbb{R}^d$: denotes the unspecified attribute space containing background and miscellaneous attributes, where $d$ is the dimensionality of the space. We enforce a $\mathbf{\mathcal{N}(\boldsymbol{0}, \boldsymbol{I})}$ prior over this space following \cite{Mathieu2016DisentanglingFO} and \cite{Jha2018DisentanglingFO}.
\end{compactitem}

Assume that we have an invertible encoding function $f$ parameterized by $\theta$, then each image $x$ can be encoded in the following way:
\[z_f, z_u = f_\theta(x), z_f \in {Z_f}, z_u \in Z_u
\]
where, we can index $z_f$ to recover the independent feature attributes \textit{i.e.} $z_f = [z_{f,1}, z_{f,2}, ..., z_{f,k}]$. To project the latent encodings back to image space, we make use of a decoding function $g$ parameterized by $\phi$. Therefore, we can get image reconstructions using the separate latent encodings.
\[\hat x = g_\phi(z_f, z_u)\]

\subsubsection{Composite Data Augmentations}

Following our initial hypothesis of recovering latent attributes using stochastic transformations, we formulate a mask-based compositional augmentation approach that leverages positive and negative transformations.

Assume that we have two sets of stochastic transformations $\mathcal{T}_p=\{p_1, p_2, ..., p_k\}$ and $\mathcal{T}_n=\{n_1, n_2, ..., n_k\}$ that can augment an image into a correlated form \textit{i.e.} $x_{aug} = t(x), t: \mathbb{R}^{C \times H \times W} \rightarrow \mathbb{R}^{C \times H \times W}, t \in \mathcal{T}_p \cup \mathcal{T}_n$. $\mathcal{T}_p$ denotes the positive set of transformations that when applied to an image should not change any of the underlying attributes, whereas $\mathcal{T}_n$ denotes the negative set of transformations that when applied to an image should change a single underlying attribute, \textit{i.e.}, when $n_i$ is applied to an image $x$, it should lead to a change only in the $z_{f,i}$ attribute and all other attributes should be preserved.

For every batch of images $x = \{x_{1}, x_{2} ... x_{B}\}$, we sample a subset of transformations to apply compositionally to $x$ and retrieve an augmented batch $x_{aug}$ and a mask vector $m \in \{0, 1\}^k$. This is further detailed in \autoref{appendix_algorithm} and \autoref{appendix_augmentation}. 

\subsubsection{Attribute Context Vectors}
\label{attribute_cv}
Taking inspiration from \cite{CPC}, we propose attribute context vectors $C^i \textcolor{white}{1} \forall i \in \{1, 2, ..., k\}$. A context vector $C^i$ is formed from each of the individual feature attributes $z_f^{i}$ through a non-linear projection. The idea is to encapsulate the batch invariant identity and variability of the $i_{th}$ attribute in $C^i$. Hence, each individual context vector should capture an independent disentangled feature space of the individual factors of variation. Assume a non-linear mapping function $\Psi : \mathbb{R}^{d \times B} \rightarrow \mathbb{R}^c$, where $c$ denotes the dimensionality of each context vector $C^i$ and $B$ denotes the size of a sampled mini-batch. We construct context vectors by aggregating all the $i^{th}$ feature attributes locally within the sampled minibatch.
\[C^i = \Psi\Big([z_{f, 1}^i, z_{f, 2}^i, ..., z_{f, B}^i]\Big) \forall i \in \{1, 2,..., k\} \]

\subsection{Loss Functions}
We describe the losses that we use to enforce disentanglement and interpretability within our feature attribute space. 
\\
We have a reconstruction penalty term to ensure that $z_u$ and $z_f$ encode enough information in order to produce high-fidelity image reconstructions.
\begin{equation}
\label{rec_loss}
    \Lagr_{R} = \sum_{i = 0}^{B} \norm{\hat{x}_{i} - x_{i}}_{2}^{2}
\end{equation}
To ensure that the unspecified attribute $z_u$ acts as a generative latent code that encodes the arbitrary features within an image, we enforce the ELBO KL objective \cite{VAE} on $z_u$.
\begin{equation}
\label{kl_loss}
    \Lagr_{KL} = KL\Big(f_{\theta}(z_u|x, z_f) \| \mathcal{N}(\boldsymbol{0}, \boldsymbol{I}))\Big)
\end{equation}
where $KL(P\|Q)=\int_{-\infty}^{\infty}p(x)\log(\frac{p(x)}{q(x)})$, $p$ and $q$ are the densities of arbitrary continuous distributions $P$ and $Q$ respectively.

We additionally enforce clustering using center loss in the feature attribute space $Z_f$ by differentiating inter-attribute features. This metric learning training strategy \cite{center_loss} promotes accumulation of feature attributes into distantly-placed clusters by providing additional self-supervision in the form of pseudo-labels obtained from the context vectors.

The center loss enforces the increment of inter-attribute distances, furthermore, diminishing the intra-attribute distances. We make use of the $\Psi$ function to project the feature attributes into the context vector space and then apply the center loss given by Equation \ref{center_loss}.
\begin{gather*}
    P_{j}^{i} = \Psi\Big([z_{f, j}^i, z_{f, j}^i, ..., z_{f, j}^i]\Big)\\ \forall i \in \{1, 2,..., k\}, j \in \{1,2,..., B\}
\end{gather*}
\begin{equation}
\label{center_loss}
    \Lagr_{cen} = \frac{1}{2} \sum_{i=1}^{k} \sum_{j=1}^{B} || P_{j}^{i} - C^{i} ||_{2}^{2}
\end{equation}
where, context vectors $C^{i}$ function as centers for the clusters corresponding to the $i^{th}$ attribute.

We also ensure that the context vectors do not deviate largely across gradient updates, by imposing a gradient descent update on the context vectors.
\begin{gather*}
C^{i} \leftarrow C^{i} - \eta\pd{\Lagr_{total}}{C^{i}}
\end{gather*}
Finally, to ensure augmentation-specific consistency within the feature attributes, we propose a feature-wise regularization penalty $\Lagr_{A}$. We first generate the augmented batch $X_{aug}$ and mask $m$ using Algorithm \ref{algoflow}. We then encode $x_{aug}$ to produce augmented feature attributes $a_f$ and unspecified attribute $a_u$ in the following way:
\begin{gather*}
a_f, a_u = f_{\theta}(x_{aug})    
\end{gather*}
Now, since we want to ensure that a specific negative augmentation $n_i$ enforces a change in only the $i_{th}$ feature attribute $z_{f}^{i}$, we encourage the representations of $a_f^{i}$ and $z_f^{i}$ to be close in representation space. Therefore, this enforces each feature attribute to be invariant to specific augmentation perturbations. Further, since the background attributes of images should be preserved irrespective of the augmentations applied, we also encourage the proximity of $z_u$ and $a_u$. This augmentation-consistency loss is defined as:

\begin{equation}
\label{augmentation_consistency_loss}
    \Lagr_{A} = \sum_{i=1}^{k}\Big(1-m[i]\Big)\norm{z_f^{i} - a_f^{i}}^{2}_{2}
    + \norm{z_u-a_u}^{2}_{2}
\end{equation}

The overall loss of our model is a weighted sum of these constituent losses.
\begin{equation}
\label{total_loss}
    \Lagr_{total} = \Lagr_R +  \lambda_{KL}\Lagr_{KL} +   \lambda_{cen}\Lagr_{cen} + \Lagr_{A}
\end{equation}

where, weights $\lambda_{KL}$ and $\lambda_{cen}$ are treated as hyperparameters. Implementation details can be found in \autoref{appendix_implementation}.

\section{Experiments}
\label{experiments}

We employ a diverse set of four datasets to evaluate the efficacy of our approach \footnote{Code is available at \href{https://github.com/sarthak268/DisCont}{https://github.com/sarthak268/DisCont}}, the details of which can be found in \autoref{appendix_dataset}.

\subsection{Quantitative Results}
\begin{figure}[H]
    \centering
    \includegraphics[scale=0.56]{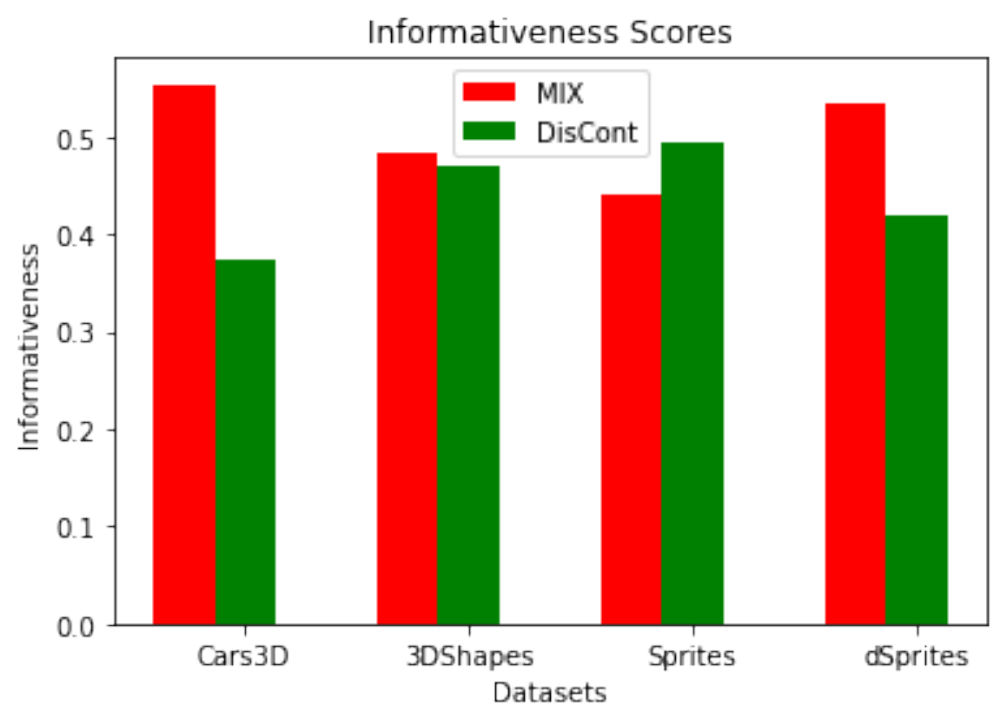}
    \vspace{-0.55cm}
    \caption{Informativeness scores for DisCont and MIX across datasets}
    \label{fig:inform_scores}
\end{figure}

\textbf{Informativeness.} To ensure a robust evaluation of our disentanglement models using unsupervised metrics, we compare informativeness scores (defined in \cite{Do2020TheoryAE}) of our model's latent chunks in Fig \ref{fig:inform_scores} with the state-of-the-art unsupervised disentanglement model presented in \cite{Hu2018DisentanglingFO} (which we refer as MIX). A lower value of informativeness score suggests a better disentangled latent representation.
Further details about the evaluation metric can be found in \autoref{appendix_informativeness}.

\subsection{Qualitative Results}
\textbf{Latent Visualization.} We present latent visualisation for the test set samples with and without the unspecified chunk, i.e. $Z_{u}$. The separation between these latent chunks in the projected space manifests the decorrelation and independence between them. Here, we present latent visualisations for the dsprites dataset, the same for the other datasets can be found in \autoref{appendix_tsne}.

\begin{figure}[H]
    \centering
    \includegraphics[width=0.5\linewidth]{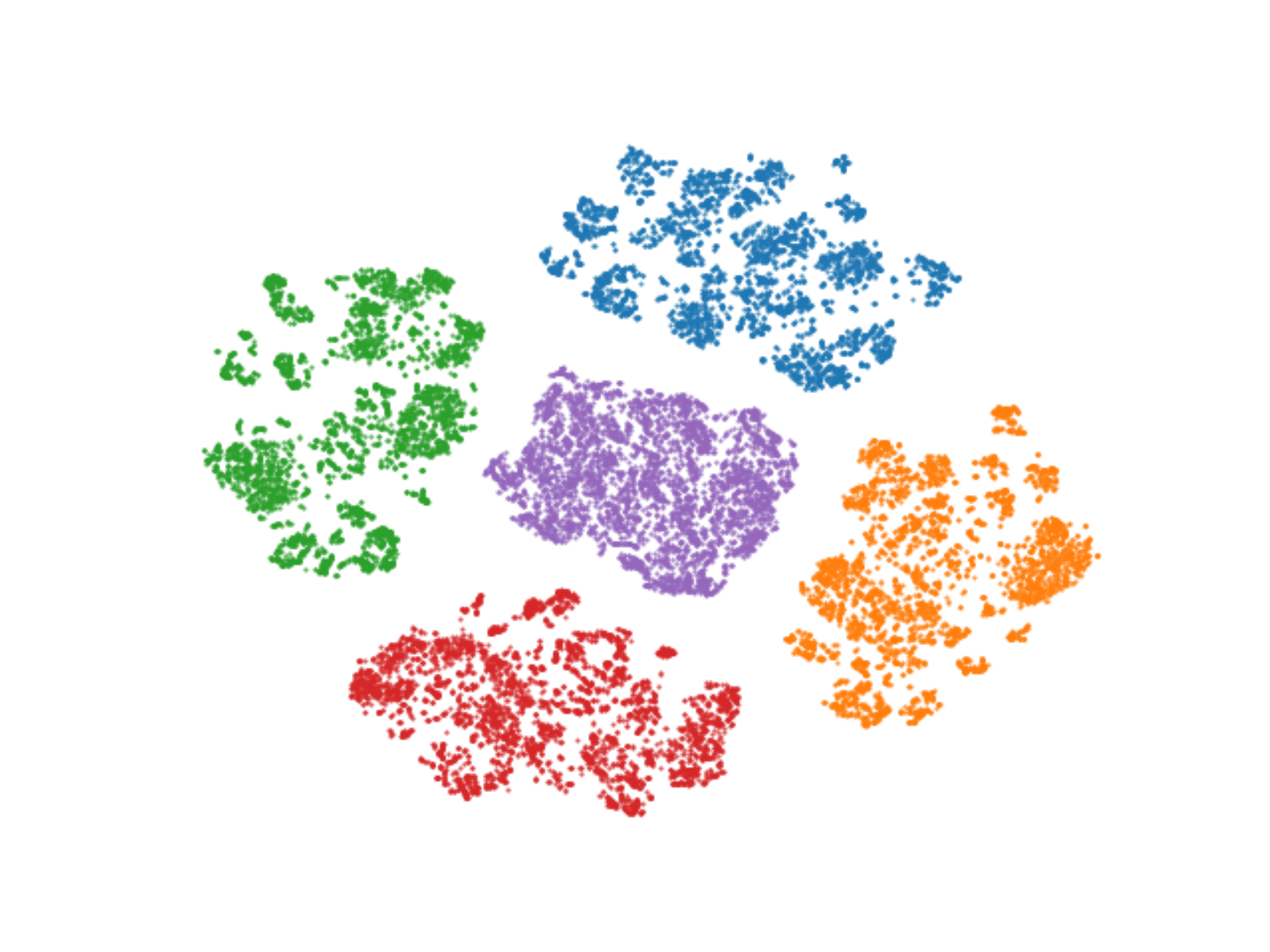}
    \includegraphics[width=0.5\linewidth]{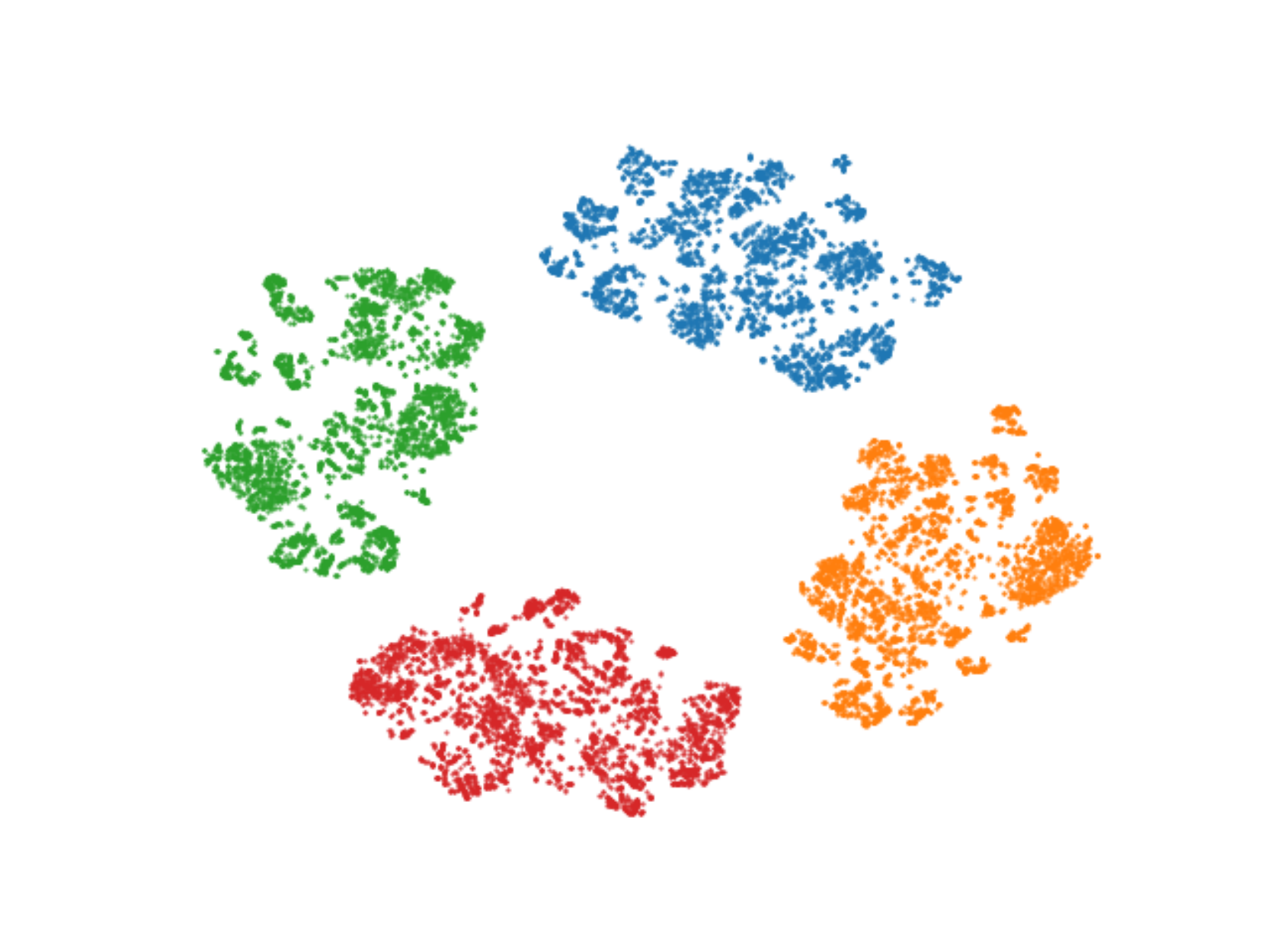}
    \caption{Latent space visualizations for dsprites dataset with all feature chunks (Top), i.e. $Z_{f} \cup Z_{u}$ and specified feature chunks (Bottom), i.e. $Z_{f}$. Each color depicts samples from different latent chunks, purple color representing the unspecified chunk, i.e. $Z_{u}$.}
    \label{fig:latent_visualisations}
\end{figure}

\textbf{Attribute Transfer.} We present attribute transfer visualizations to construe the quality of disentanglement. The images in the first two rows in each grid are randomly sampled from the test-set. The bottom row images are formed by swapping one specified attribute from the top row image with the corresponding attribute chunk in the second row image, keep all other attributes fixed. 
This allows us to quantify the purity of attribute-wise information captured by each latent chunk. 
We present these results for Cars3D and 3DShapes here, while the others in \autoref{attribute_transfer_appendix}.

\begin{figure}[H]
    \centering
    \subfloat[Cars3D; Specified Attribute: Color]{
    \includegraphics[width=0.8\linewidth]{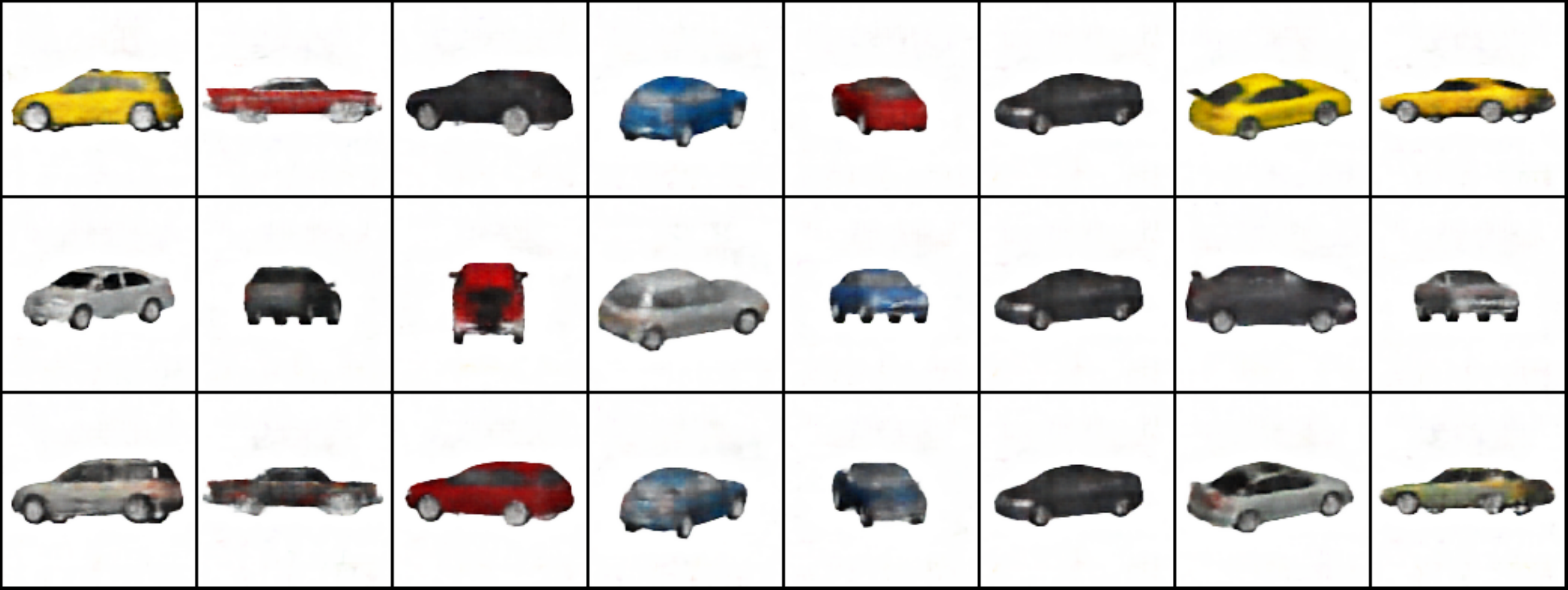}}\\
    \subfloat[3DShapes; Specified Attribute: Orientation]{
    \includegraphics[width=0.8\linewidth]{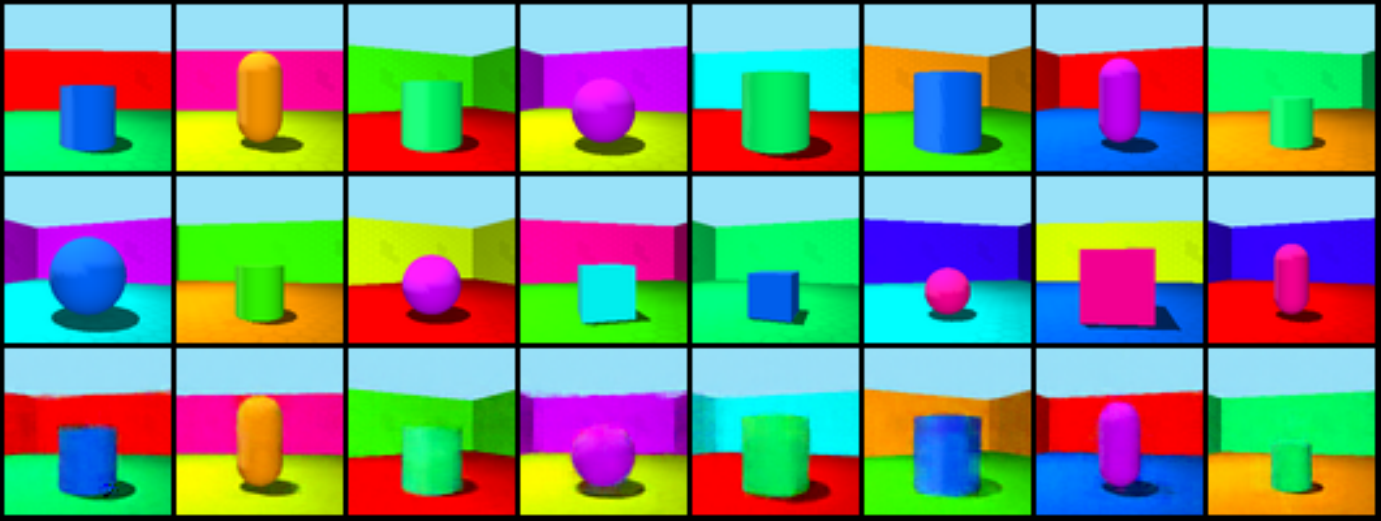}}
    \caption{Attribute transfer results obtained by swapping the specified chunk.}
    \label{fig:style_transfer}
\end{figure}

\section{Conclusion}
\label{conclusion}
In this paper, we propose a self-supervised attribute disentanglement framework \textit{DisCont} which leverages specific data augmentations to exploit the spatial inductive biases present in images. We also propose `Attribute Context Vectors' that encapsulate the intra-attribute variations. Our results show that such a framework can be readily used to recover semantically meaningful attributes independently. 

\section*{\fontsize{10}{10}\selectfont Acknowledgement} \hspace{-0.1cm} We would like to thank Dr. Saket Anand (IIIT Delhi) for his guidance in formulating the initial problem statement, and valuable comments and feedback on this paper. We would also like to thank Ananya Harsh Jha (NYU) for providing the initial ideas.

\bibliography{main}

\begin{thebibliography}{48}
\providecommand{\natexlab}[1]{#1}
\providecommand{\url}[1]{\texttt{#1}}
\expandafter\ifx\csname urlstyle\endcsname\relax
  \providecommand{\doi}[1]{doi: #1}\else
  \providecommand{\doi}{doi: \begingroup \urlstyle{rm}\Url}\fi

\bibitem[Arora et~al.(2019)Arora, Khandeparkar, Khodak, Plevrakis, and
  Saunshi]{theoretical_contrastive}
Arora, S., Khandeparkar, H., Khodak, M., Plevrakis, O., and Saunshi, N.
\newblock A theoretical analysis of contrastive unsupervised representation
  learning.
\newblock \emph{arXiv preprint arXiv:1902.09229}, 2019.

\bibitem[Bengio(2013)]{deep_rep}
Bengio, Y.
\newblock Deep learning of representations: Looking forward.
\newblock In \emph{International Conference on Statistical Language and Speech
  Processing}, pp.\  1--37. Springer, 2013.

\bibitem[Bengio et~al.(2013)Bengio, Courville, and
  Vincent]{bengio2013representation}
Bengio, Y., Courville, A., and Vincent, P.
\newblock Representation learning: A review and new perspectives.
\newblock \emph{IEEE transactions on pattern analysis and machine
  intelligence}, 35\penalty0 (8):\penalty0 1798--1828, 2013.

\bibitem[Bhagat et~al.(2020)Bhagat, Uppal, Yin, and
  Lim]{Bhagat2020DisentanglingRU}
Bhagat, S., Uppal, S., Yin, V.~T., and Lim, N.
\newblock Disentangling representations using gaussian processes in variational
  autoencoders for video prediction.
\newblock \emph{ArXiv}, abs/2001.02408, 2020.

\bibitem[Burgess \& Kim(2018)Burgess and Kim]{3dshapes18}
Burgess, C. and Kim, H.
\newblock 3d shapes dataset.
\newblock https://github.com/deepmind/3dshapes-dataset/, 2018.

\bibitem[Cao et~al.(2018)Cao, Katzir, Jiang, Lischinski, Cohen-Or, Tu, and
  Li]{Cao2018DiDADS}
Cao, J., Katzir, O., Jiang, P., Lischinski, D., Cohen-Or, D., Tu, C., and Li,
  Y.
\newblock Dida: Disentangled synthesis for domain adaptation.
\newblock \emph{ArXiv}, abs/1805.08019, 2018.

\bibitem[Chen et~al.(2020)Chen, Kornblith, Norouzi, and Hinton]{SimCLR}
Chen, T., Kornblith, S., Norouzi, M., and Hinton, G.
\newblock A simple framework for contrastive learning of visual
  representations, 2020.

\bibitem[Chen et~al.(2018)Chen, Li, Grosse, and Duvenaud]{Chen2018IsolatingSO}
Chen, T.~Q., Li, X., Grosse, R.~B., and Duvenaud, D.
\newblock Isolating sources of disentanglement in variational autoencoders.
\newblock In \emph{NeurIPS}, 2018.

\bibitem[Chen et~al.(2016)Chen, Duan, Houthooft, Schulman, Sutskever, and
  Abbeel]{infogan}
Chen, X., Duan, Y., Houthooft, R., Schulman, J., Sutskever, I., and Abbeel, P.
\newblock Infogan: Interpretable representation learning by information
  maximizing generative adversarial nets.
\newblock In \emph{Advances in neural information processing systems}, pp.\
  2172--2180, 2016.

\bibitem[Cohen \& Shashua(2016)Cohen and Shashua]{inductive_bias_1}
Cohen, N. and Shashua, A.
\newblock Inductive bias of deep convolutional networks through pooling
  geometry.
\newblock \emph{arXiv preprint arXiv:1605.06743}, 2016.

\bibitem[Denton \& Birodkar(2017)Denton and Birodkar]{Denton2017UnsupervisedLO}
Denton, E.~L. and Birodkar, V.
\newblock Unsupervised learning of disentangled representations from video.
\newblock In \emph{NIPS}, 2017.

\bibitem[Do \& Tran(2020)Do and Tran]{Do2020TheoryAE}
Do, K. and Tran, T.
\newblock Theory and evaluation metrics for learning disentangled
  representations.
\newblock \emph{ArXiv}, abs/1908.09961, 2020.

\bibitem[{Farhadi} et~al.(2009){Farhadi}, {Endres}, {Hoiem}, and
  {Forsyth}]{describing_attributes}
{Farhadi}, A., {Endres}, I., {Hoiem}, D., and {Forsyth}, D.
\newblock Describing objects by their attributes.
\newblock In \emph{2009 IEEE Conference on Computer Vision and Pattern
  Recognition}, pp.\  1778--1785, 2009.

\bibitem[Ghosh \& Gupta(2019)Ghosh and Gupta]{inductive_bias_2}
Ghosh, R. and Gupta, A.~K.
\newblock Investigating convolutional neural networks using spatial orderness.
\newblock In \emph{Proceedings of the IEEE International Conference on Computer
  Vision Workshops}, pp.\  0--0, 2019.

\bibitem[Gutmann \& Hyv{\"a}rinen(2010)Gutmann and Hyv{\"a}rinen]{NCE}
Gutmann, M. and Hyv{\"a}rinen, A.
\newblock Noise-contrastive estimation: A new estimation principle for
  unnormalized statistical models.
\newblock In \emph{Proceedings of the Thirteenth International Conference on
  Artificial Intelligence and Statistics}, pp.\  297--304, 2010.

\bibitem[He et~al.(2019)He, Fan, Wu, Xie, and Girshick]{moco}
He, K., Fan, H., Wu, Y., Xie, S., and Girshick, R.
\newblock Momentum contrast for unsupervised visual representation learning.
\newblock \emph{arXiv preprint arXiv:1911.05722}, 2019.

\bibitem[Higgins et~al.(2017)Higgins, Matthey, Pal, Burgess, Glorot, Botvinick,
  Mohamed, and Lerchner]{Higgins2017betaVAELB}
Higgins, I., Matthey, L., Pal, A., Burgess, C., Glorot, X., Botvinick, M.~M.,
  Mohamed, S., and Lerchner, A.
\newblock beta-vae: Learning basic visual concepts with a constrained
  variational framework.
\newblock In \emph{ICLR}, 2017.

\bibitem[Hsieh et~al.(2018)Hsieh, Liu, Huang, Fei-Fei, and
  Niebles]{Hsieh2018LearningTD}
Hsieh, J.-T., Liu, B., Huang, D.-A., Fei-Fei, L., and Niebles, J.~C.
\newblock Learning to decompose and disentangle representations for video
  prediction.
\newblock In \emph{NeurIPS}, 2018.

\bibitem[Hu et~al.(2018)Hu, Szab{\'o}, Portenier, Zwicker, and
  Favaro]{Hu2018DisentanglingFO}
Hu, Q., Szab{\'o}, A., Portenier, T., Zwicker, M., and Favaro, P.
\newblock Disentangling factors of variation by mixing them.
\newblock \emph{2018 IEEE/CVF Conference on Computer Vision and Pattern
  Recognition}, pp.\  3399--3407, 2018.

\bibitem[Hénaff et~al.(2019)Hénaff, Srinivas, Fauw, Razavi, Doersch, Eslami,
  and van~den Oord]{CPC2}
Hénaff, O.~J., Srinivas, A., Fauw, J.~D., Razavi, A., Doersch, C., Eslami, S.
  M.~A., and van~den Oord, A.
\newblock Data-efficient image recognition with contrastive predictive coding,
  2019.

\bibitem[Jha et~al.(2018)Jha, Anand, Singh, and
  Veeravasarapu]{Jha2018DisentanglingFO}
Jha, A.~H., Anand, S., Singh, M.~K., and Veeravasarapu, V. S.~R.
\newblock Disentangling factors of variation with cycle-consistent variational
  auto-encoders.
\newblock \emph{ArXiv}, abs/1804.10469, 2018.

\bibitem[Kim \& Mnih(2018)Kim and Mnih]{Kim2018DisentanglingBF}
Kim, H. and Mnih, A.
\newblock Disentangling by factorising.
\newblock \emph{ArXiv}, abs/1802.05983, 2018.

\bibitem[Kim et~al.(2019)Kim, Wang, Sahu, and Pavlovic]{Kim2019RelevanceFV}
Kim, M., Wang, Y., Sahu, P., and Pavlovic, V.
\newblock Relevance factor vae: Learning and identifying disentangled factors.
\newblock \emph{ArXiv}, abs/1902.01568, 2019.

\bibitem[Kingma \& Welling(2013)Kingma and Welling]{VAE}
Kingma, D.~P. and Welling, M.
\newblock Auto-encoding variational bayes.
\newblock \emph{arXiv preprint arXiv:1312.6114}, 2013.

\bibitem[Kipf et~al.(2019)Kipf, van~der Pol, and
  Welling]{structured_world_models}
Kipf, T., van~der Pol, E., and Welling, M.
\newblock Contrastive learning of structured world models.
\newblock \emph{arXiv preprint arXiv:1911.12247}, 2019.

\bibitem[Le \& Akoglu(2019)Le and Akoglu]{contravis}
Le, T. and Akoglu, L.
\newblock Contravis: contrastive and visual topic modeling for comparing
  document collections.
\newblock In \emph{The World Wide Web Conference}, pp.\  928--938, 2019.

\bibitem[Liu et~al.(2015)Liu, Luo, Wang, and Tang]{celeba}
Liu, Z., Luo, P., Wang, X., and Tang, X.
\newblock Deep learning face attributes in the wild.
\newblock In \emph{Proceedings of International Conference on Computer Vision
  (ICCV)}, December 2015.

\bibitem[Locatello et~al.(2018)Locatello, Bauer, Lucic, R{\"a}tsch, Gelly,
  Sch{\"o}lkopf, and Bachem]{loca}
Locatello, F., Bauer, S., Lucic, M., R{\"a}tsch, G., Gelly, S., Sch{\"o}lkopf,
  B., and Bachem, O.
\newblock Challenging common assumptions in the unsupervised learning of
  disentangled representations.
\newblock \emph{arXiv preprint arXiv:1811.12359}, 2018.

\bibitem[Ma et~al.(2019)Ma, Zhou, Cui, Yang, and Zhu]{Ma2019LearningDR}
Ma, J., Zhou, C., Cui, P., Yang, H., and Zhu, W.
\newblock Learning disentangled representations for recommendation.
\newblock In \emph{NeurIPS}, 2019.

\bibitem[Mathieu et~al.(2016)Mathieu, Zhao, Sprechmann, Ramesh, and
  LeCun]{Mathieu2016DisentanglingFO}
Mathieu, M., Zhao, J.~J., Sprechmann, P., Ramesh, A., and LeCun, Y.
\newblock Disentangling factors of variation in deep representation using
  adversarial training.
\newblock \emph{ArXiv}, abs/1611.03383, 2016.

\bibitem[Matthey et~al.(2017)Matthey, Higgins, Hassabis, and
  Lerchner]{dsprites17}
Matthey, L., Higgins, I., Hassabis, D., and Lerchner, A.
\newblock dsprites: Disentanglement testing sprites dataset.
\newblock https://github.com/deepmind/dsprites-dataset/, 2017.

\bibitem[Meng et~al.(2019)Meng, Pawlowski, Rueckert, and
  Kainz]{Meng2019RepresentationDF}
Meng, Q., Pawlowski, N., Rueckert, D., and Kainz, B.
\newblock Representation disentanglement for multi-task learning with
  application to fetal ultrasound.
\newblock In \emph{SUSI/PIPPI@MICCAI}, 2019.

\bibitem[Misra \& van~der Maaten(2019)Misra and van~der Maaten]{PIRL}
Misra, I. and van~der Maaten, L.
\newblock Self-supervised learning of pretext-invariant representations.
\newblock \emph{arXiv preprint arXiv:1912.01991}, 2019.

\bibitem[Patrick et~al.(2020)Patrick, Asano, Fong, Henriques, Zweig, and
  Vedaldi]{multi-modal-contrast}
Patrick, M., Asano, Y.~M., Fong, R., Henriques, J.~F., Zweig, G., and Vedaldi,
  A.
\newblock Multi-modal self-supervision from generalized data transformations.
\newblock \emph{arXiv preprint arXiv:2003.04298}, 2020.

\bibitem[Peters et~al.(2017)Peters, Janzing, and
  Sch{\"o}lkopf]{peters2017elements}
Peters, J., Janzing, D., and Sch{\"o}lkopf, B.
\newblock \emph{Elements of causal inference: foundations and learning
  algorithms}.
\newblock MIT press, 2017.

\bibitem[Reed et~al.(2015)Reed, Zhang, Zhang, and Lee]{Reed2015DeepVA}
Reed, S.~E., Zhang, Y., Zhang, Y., and Lee, H.
\newblock Deep visual analogy-making.
\newblock In \emph{NIPS}, 2015.

\bibitem[Sch{\"o}lkopf et~al.(2012)Sch{\"o}lkopf, Janzing, Peters, Sgouritsa,
  Zhang, and Mooij]{causal}
Sch{\"o}lkopf, B., Janzing, D., Peters, J., Sgouritsa, E., Zhang, K., and
  Mooij, J.
\newblock On causal and anticausal learning.
\newblock \emph{arXiv preprint arXiv:1206.6471}, 2012.

\bibitem[Shi et~al.(2019)Shi, Liang, Hou, Li, Liu, and Zhang]{deepchannel}
Shi, J., Liang, C., Hou, L., Li, J., Liu, Z., and Zhang, H.
\newblock Deepchannel: Salience estimation by contrastive learning for
  extractive document summarization.
\newblock In \emph{Proceedings of the AAAI Conference on Artificial
  Intelligence}, volume~33, pp.\  6999--7006, 2019.

\bibitem[Shukla et~al.(2019)Shukla, Bhagat, Uppal, Anand, and
  Turaga]{Shukla2019ProductOO}
Shukla, A., Bhagat, S., Uppal, S., Anand, S., and Turaga, P.~K.
\newblock Product of orthogonal spheres parameterization for disentangled
  representation learning.
\newblock In \emph{BMVC}, 2019.

\bibitem[Srinivas et~al.(2020)Srinivas, Laskin, and Abbeel]{curl}
Srinivas, A., Laskin, M., and Abbeel, P.
\newblock Curl: Contrastive unsupervised representations for reinforcement
  learning.
\newblock \emph{arXiv preprint arXiv:2004.04136}, 2020.

\bibitem[Tian et~al.(2019)Tian, Krishnan, and Isola]{CMC}
Tian, Y., Krishnan, D., and Isola, P.
\newblock Contrastive multiview coding, 2019.

\bibitem[Udandarao et~al.(2020)Udandarao, Maiti, Srivatsav, Vyalla, Yin, and
  Shah]{cobra}
Udandarao, V., Maiti, A., Srivatsav, D., Vyalla, S.~R., Yin, Y., and Shah,
  R.~R.
\newblock Cobra: Contrastive bi-modal representation algorithm.
\newblock \emph{arXiv preprint arXiv:2005.03687}, 2020.

\bibitem[van~den Oord et~al.(2018)van~den Oord, Li, and Vinyals]{CPC}
van~den Oord, A., Li, Y., and Vinyals, O.
\newblock Representation learning with contrastive predictive coding, 2018.

\bibitem[Villegas et~al.(2017)Villegas, Yang, Hong, Lin, and
  Lee]{Villegas2017DecomposingMA}
Villegas, R., Yang, J., Hong, S., Lin, X., and Lee, H.
\newblock Decomposing motion and content for natural video sequence prediction.
\newblock \emph{ArXiv}, abs/1706.08033, 2017.

\bibitem[Vu \& Huang(2019)Vu and Huang]{Vu2019DomainAM}
Vu, H.~T. and Huang, C.-C.
\newblock Domain adaptation meets disentangled representation learning and
  style transfer.
\newblock \emph{2019 IEEE International Conference on Systems, Man and
  Cybernetics (SMC)}, pp.\  2998--3005, 2019.

\bibitem[Wen et~al.(2016)Wen, Zhang, Li, and Qiao]{center_loss}
Wen, Y., Zhang, K., Li, Z., and Qiao, Y.
\newblock A discriminative feature learning approach for deep face recognition.
\newblock In \emph{European conference on computer vision}, pp.\  499--515.
  Springer, 2016.

\bibitem[Yang et~al.(2019)Yang, Dvornek, Zhang, Chapiro, Lin, and
  Duncan]{Yang2019UnsupervisedDA}
Yang, J., Dvornek, N.~C., Zhang, F., Chapiro, J., Lin, M., and Duncan, J.~S.
\newblock Unsupervised domain adaptation via disentangled representations:
  Application to cross-modality liver segmentation.
\newblock \emph{MICCAI}, 11765:\penalty0 255--263, 2019.

\bibitem[Zhang et~al.(2019)Zhang, Huang, Li, Zhao, and Zhang]{zhang2019multi}
Zhang, J., Huang, Y., Li, Y., Zhao, W., and Zhang, L.
\newblock Multi-attribute transfer via disentangled representation.
\newblock In \emph{Proceedings of the AAAI Conference on Artificial
  Intelligence}, volume~33, pp.\  9195--9202, 2019.

\end{thebibliography}
\bibliographystyle{icml2020}
\newpage
\appendix
\onecolumn
\section{Mask and Augmented Batch Generation Algorithm}
This section describes the generation of the mask and the augmented batch for the computation of the augmentation-consistency loss $\Lagr_{A}$ (refer \autoref{augmentation_consistency_loss}). The entire algorithm is detailed below:
\label{appendix_algorithm}

{\centering
\centerline{
\begin{minipage}{.6\linewidth}
\begin{algorithm}[H]
  \caption{Mask and Augmented Batch generation}
  \label{algoflow}
\centering
\begin{algorithmic}
  \STATE {\bfseries Input:} A batch of images $x$, the set of positive transformations $\mathcal{T}_p$, the set of negative transformations $\mathcal{T}_n$, number of feature attributes $k$
  \STATE {\bfseries Output:} The augmented batch $x_{aug}$, the mask $m$
  \STATE Initialize $m=[0, 0,..., 0]_k$, $x_{aug}=x$
  \FOR{$i=1$ {\bfseries to} $k$}
  \STATE $p \sim Bernoulli(0.5)$
  \IF{$p=1$} 
  \STATE $m[i]=1$
  \STATE $x_{aug}=n_i(x_{aug})$
  \ENDIF
  \ENDFOR
      \FOR{$i=1$ {\bfseries to} $k$}
  \STATE $p \sim Bernoulli(0.5)$
  \IF{$p=1$} 
  \STATE $x_{aug}=p_i(x_{aug})$
  \ENDIF
  \ENDFOR
  \STATE return $x_{aug}, m$
\end{algorithmic}
\end{algorithm}
\end{minipage}}
}

\section{Augmentations}
\label{appendix_augmentation}
The set of augmentations and their sampling range of parameters that we use in this work are detailed in the table below:
\begin{table}[H]
    \centering
    \begin{tabular}{ccc}
        \toprule
        & \textbf{Positive Augmentations $\mathcal{T}_p$} &\\
        \midrule
        \textit{Type} & \textit{Parameter} & \textit{Range}\\
        \midrule
        Gaussian Noise & $\sigma$ & [0.5, 1, 2, 5]\\
        Gaussian Smoothing & $\sigma$ & [0.1, 0.2, 0.5, 1]\\
        \midrule
        & \textbf{Negative Augmentations $\mathcal{T}_n$} &\\
        \midrule
        \textit{Type} & \textit{Parameter} & \textit{Range}\\
        \midrule
        Grayscale Transform & -- & --\\
        Flipping & Orientation & [Horizontal, Vertical]\\
        Rotation & $\theta$ & [90\degree, 180\degree, 270\degree]\\
        Crop \& Resize & -- & --\\
        Cutout & Length & [5, 10, 15, 20]\\
        \bottomrule
    \end{tabular}
    \caption{Set of augmentations used for training \textit{DisCont}.}
    \label{tab:augs}
\end{table}
\newpage

\section{Dataset Description}
\label{appendix_dataset}
We use the following datasets to evaluate our model performance:
\begin{compactitem}
    \setlength{\itemsep}{1pt}
    \setlength{\parskip}{0pt}
    \setlength{\parsep}{0pt}
    \item \textbf{Sprites} \cite{Reed2015DeepVA} is a dataset of 480 unique animated caricatures (sprites) with essentially 6 factors of variations namely gender, hair type, body type, armor type, arm type and greaves type. The entire dataset consists of 143,040 images with 320, 80 and 80 characters in train, test and validation sets respectively.
  \item \textbf{Cars3D} \cite{Reed2015DeepVA} consists of a total of 17,568 images of synthetic cars with elevation, azimuth and object type varying in each image.
  \item \textbf{3DShapes} \cite{3dshapes18} is a dataset of 3D shapes generated using 6 independent factors of variations namely floor color, wall color, object color, scale, shape and orientation. It consists of a total of 480,000 images of 4 discrete 3D shapes.
  \item \textbf{dSprites} \cite{dsprites17} is a dataset consisting of 2D square, ellipse, and hearts with varying color, scale, rotation, x and y positions. In total, it consists of 737,280 gray-scale images.
\end{compactitem}

\section{Implementation, Training and Hyperparameter Details}
\label{appendix_implementation}
We use a single experimental setup across all our experiments. We implement all our models on an Nvidia GTX 1080 GPU using the PyTorch framework. Architectural details are provided in \autoref{tab:implementation}. The training hyperparameters used are listed in \autoref{tab:hyperparameters_}. 

\begin{table}[H]
    \centering
  \begin{tabular}{ccc}
    \toprule
    \textbf{Encoder $f$}&\textbf{Decoder $g$}&\textbf{Context Network $\Psi$} \\
    \midrule
    Input: $\mathbb{R}^{64\times64\times3}$& Input: $\mathbb{R}^{(k+1) \times d}$& Input: $\mathbb{R}^{k \times d}$\\
    Conv $4\times4$, 64, ELU, stride 2, BN& FC, $(k+1) \times d$, 1024, ReLU, BN & FC, $k \times d$, 4096, ReLU\\
    Conv $3\times3$, 128, ELU, stride 2, BN& FC, 1024, 4068, ReLU, BN & FC, 4096, $k \times c$, ReLU\\
    Conv $3\times3$, 256, ELU, stride 2, BN& Deconv, $3\times3$, 256, ReLU, stride 2, BN & \\
    Conv $3\times3$, 512, ELU, stride 2, BN&Deconv, $3\times3$, 128, ReLU, stride 2, BN &\\
    FC, 4608, 1024, ELU, BN&Deconv, $3\times3$, 64, ReLU, stride 2, BN &\\
    FC, 1024, $(k+1) \times d$, ELU, BN&Deconv, $4\times4$, 3, ReLU, stride 2, BN &\\
  \bottomrule
\end{tabular}
\caption{Architectures for encoder $f$, decoder $g$ and context network $\Psi$ for all experiments. Here, Conv denotes 2D convolution layer, Deconv denotes 2D transposed convolution layer, FC denotes fully connected layer, BN denotes batch normalisation layer.}
  \label{tab:implementation}
\end{table}

\begin{table}[H]
    \centering
    \begin{tabular}{cc}
        \toprule
        \textbf{Parameter} & \textbf{Value} \\
        \midrule
        Batch Size ($B$) & 64\\
        Latent Space Dimension ($d$) & 32\\ 
        Number of Feature Attributes ($k$) & 2\\
        Context Vector Dimension ($c$) & 100\\
        KL-Divergence Weight ($\lambda_{KL}$) & 1\\
        Augmentation-Consistency Loss Weight ($\lambda_A$) & 0.2\\
         Optimizer & Adam\\
         Learning Rate ($\eta$) & 1e-4\\
         Adam: $\beta_1$ & 0.5\\
         Adam: $\beta_2$ & 0.999\\
         Training Epochs & 250\\
        \bottomrule
    \end{tabular}
    \caption{Training hyperparameters for \textit{DisCont} common across all datasets.}
    \label{tab:hyperparameters_}
\end{table}

\newpage

\section{Informativeness Evaluation Metric}
\label{appendix_informativeness}
As detailed in \cite{Do2020TheoryAE}, disentangled representations need to have low mutual information with the base data distribution, since ideally each representation should capture atmost one attribute within the data. The informativeness of a representation $z_{f,i}$ \textit{w.r.t} data $x$ is determined by computing the mutual information $\mathbb{I}(x, z_{f,i})$ using the following Equation:

\begin{gather}
    \mathbb{I}(x, z_{f,i}) = \int_{x} \int_{z_{f}} p_{\mathcal{D}}(x)f(z_{f,i}|x)\log\Bigg(\frac{f(z_{f,i}|x)}{f(z_{f,i})}\Bigg) \,dx\,dz_{f}
\end{gather}
where $f(\cdot)$ depicts the encoding function, i.e., $f: \mathbb{R}^{C \times H \times W} \rightarrow \mathbb{R}^{d}$ and $\mathcal{D}$ depicts the dataset of all samples, such that $\mathcal{D} = \{x_{i}\}_{i=1}^{N}$.
The informativeness metric helps us capture the amount of information encapsulated within each latent chunk with respect to the original image $x$. We compare our \textit{DisCont} model with unsupervised disentanglement model proposed by \cite{Hu2018DisentanglingFO}. For training the model in \cite{Hu2018DisentanglingFO}, we use the following hyperparameter values.

\begin{table}[H]
    \centering
    \begin{tabular}{ccccc}
        \toprule
            \textbf{Hyperparameter} & \textbf{Sprites} & \textbf{Cars3D} & \textbf{3DShapes} & \textbf{dSprites} \\
            \midrule
            Latent Space Dimension ($d$) & 256 & 96 & 512 & 96 \\ 
            Number of Chunks & 8 & 3 & 8 & 3 \\
            Dimension of Each Chunk & 32 & 32 & 64 & 32 \\
            Optimizer & Adam & Adam & Adam & Adam \\ 
            Learning Rate ($\eta$) & 2e-4 & 2e-4 & 5e-5 & 2e-4 \\ 
            Adam: $\beta_{1}$ & 0.5 & 0.5 & 0.5 & 0.5 \\
            Adam: $\beta_{1}$ & 0.99 & 0.99 & 0.99 & 0.99 \\
            Training Epochs & 100 & 200 & 200 & 150 \\
        \bottomrule
    \end{tabular}
    \caption{Training hyperparameters for the model in \cite{Hu2018DisentanglingFO}}
    \label{tab:hyperparameters}
\end{table}

\section{Attribute Transfer}
\label{attribute_transfer_appendix}

\begin{figure}[H]
    \centering
    \subfloat[Specified Attribute: Hair Color]{
    \includegraphics[width=0.4\linewidth]{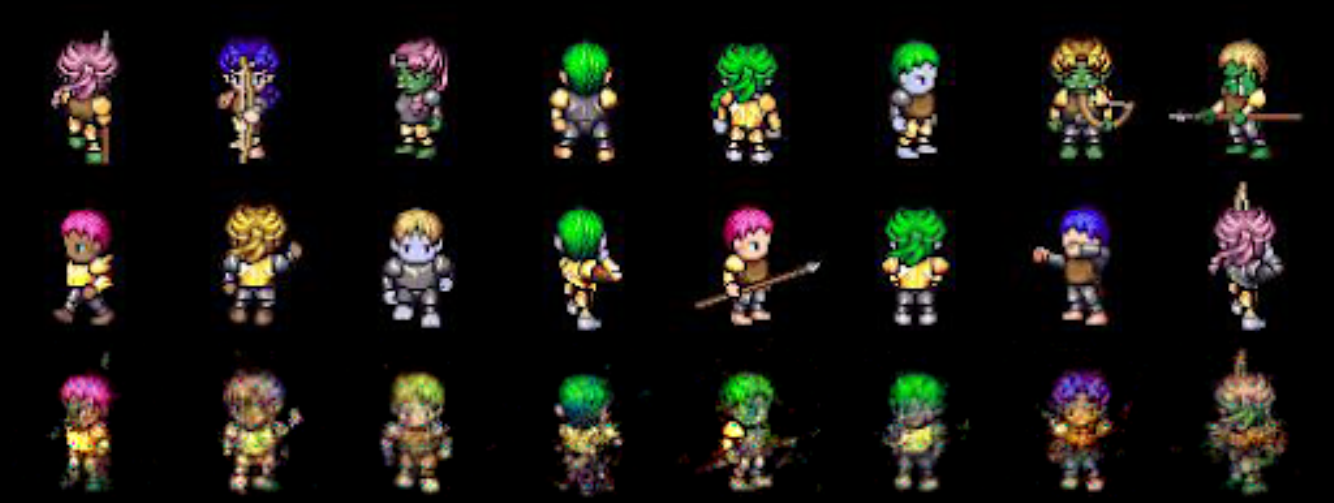}}
    \subfloat[Specified Attribute: Pose]{
    \includegraphics[width=0.4\linewidth]{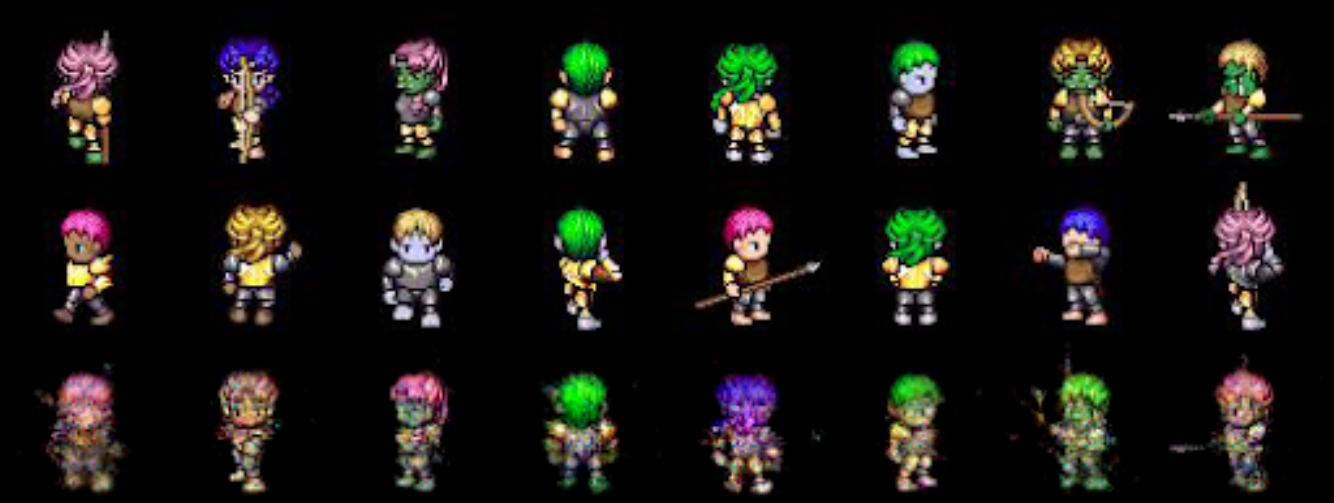}}
    \caption{Attribute transfer results for \textit{DisCont} obtained by swapping the specified chunk.}
    \label{fig:style_transfer_app}
\end{figure}

We present attribute transfer visualizations for validating our disentanglement performance. 
The first two rows depict the sampled batch of images from the test set while the bottom row depicts the images generated by swapping the specified attribute from the first row images with that of the second row images. The style transfer results for Sprites are shown in Fig \ref{fig:style_transfer_app}. The feature swapping results for the dSprites dataset were not consistent probably because of the ambiguity induced by the color transformation in the feature attribute space, when applied to the single channeled images.

\section{Latent Visualisation}
\label{appendix_tsne}
\begin{figure}[H]
    \centering
    \subfloat[Cars3D]{
        \includegraphics[width=0.3\linewidth]{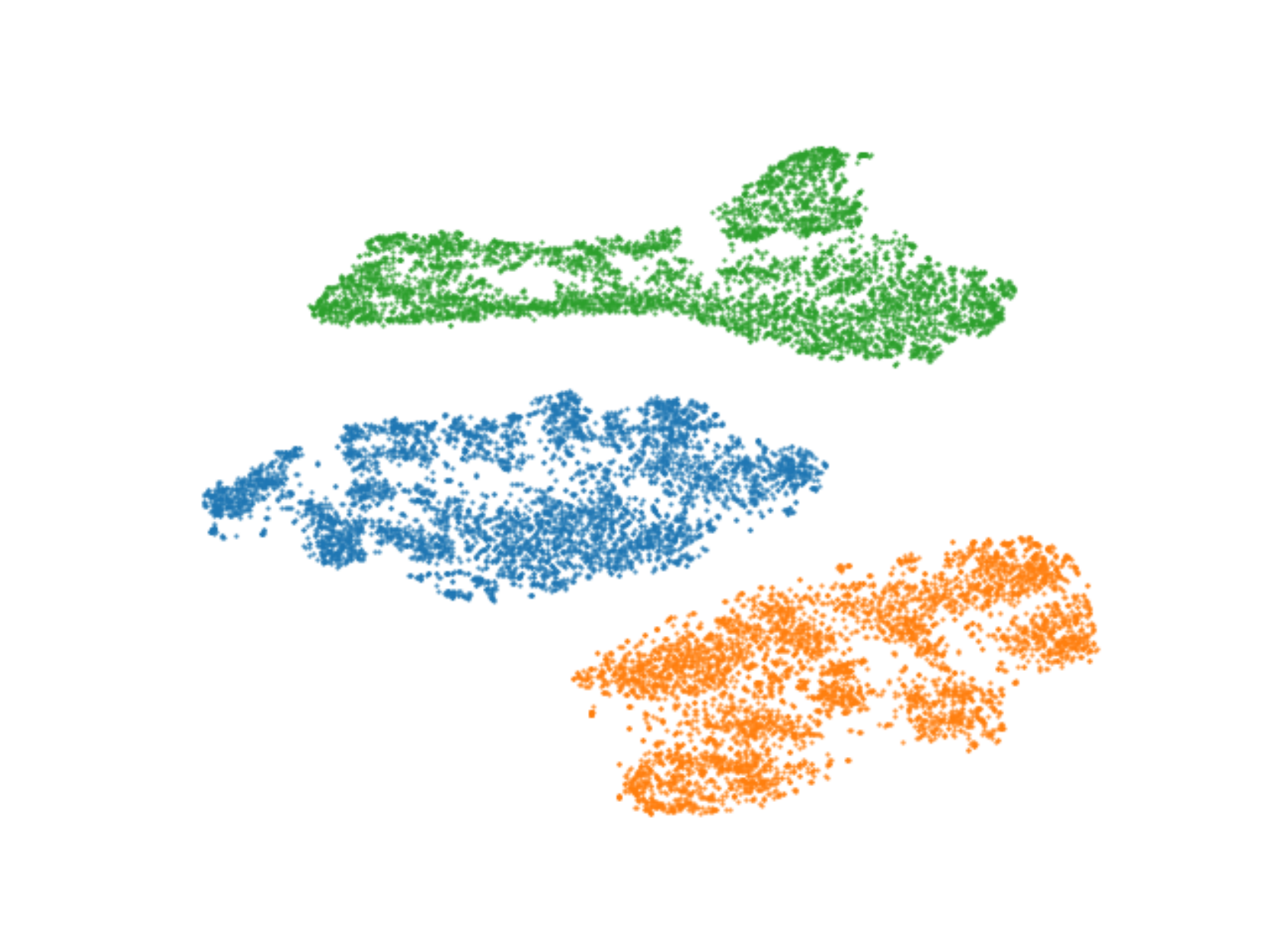}
        \includegraphics[width=0.3\linewidth]{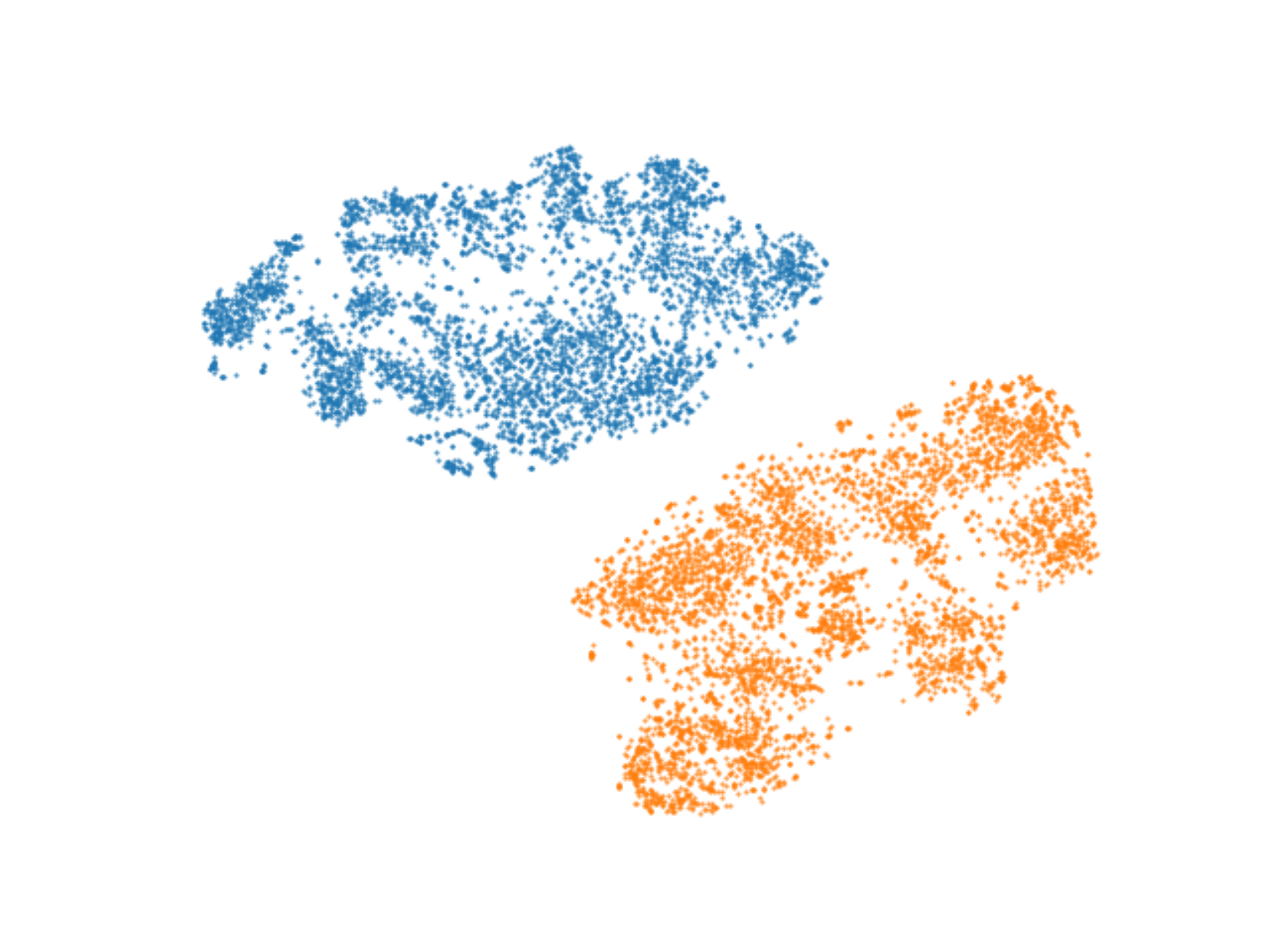}}\\
    \subfloat[3DShapes]{
        \includegraphics[width=0.3\linewidth]{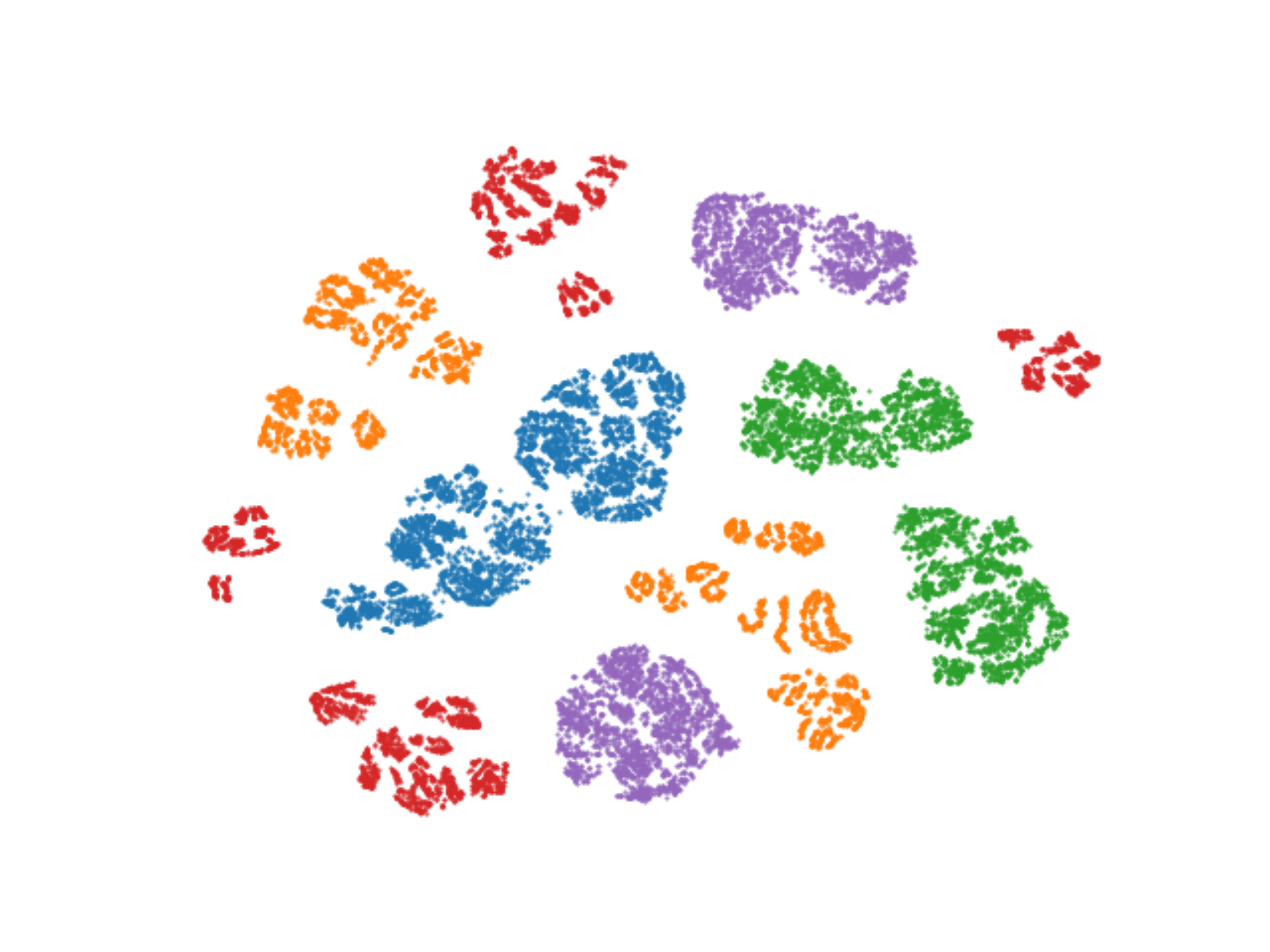}
        \includegraphics[width=0.3\linewidth]{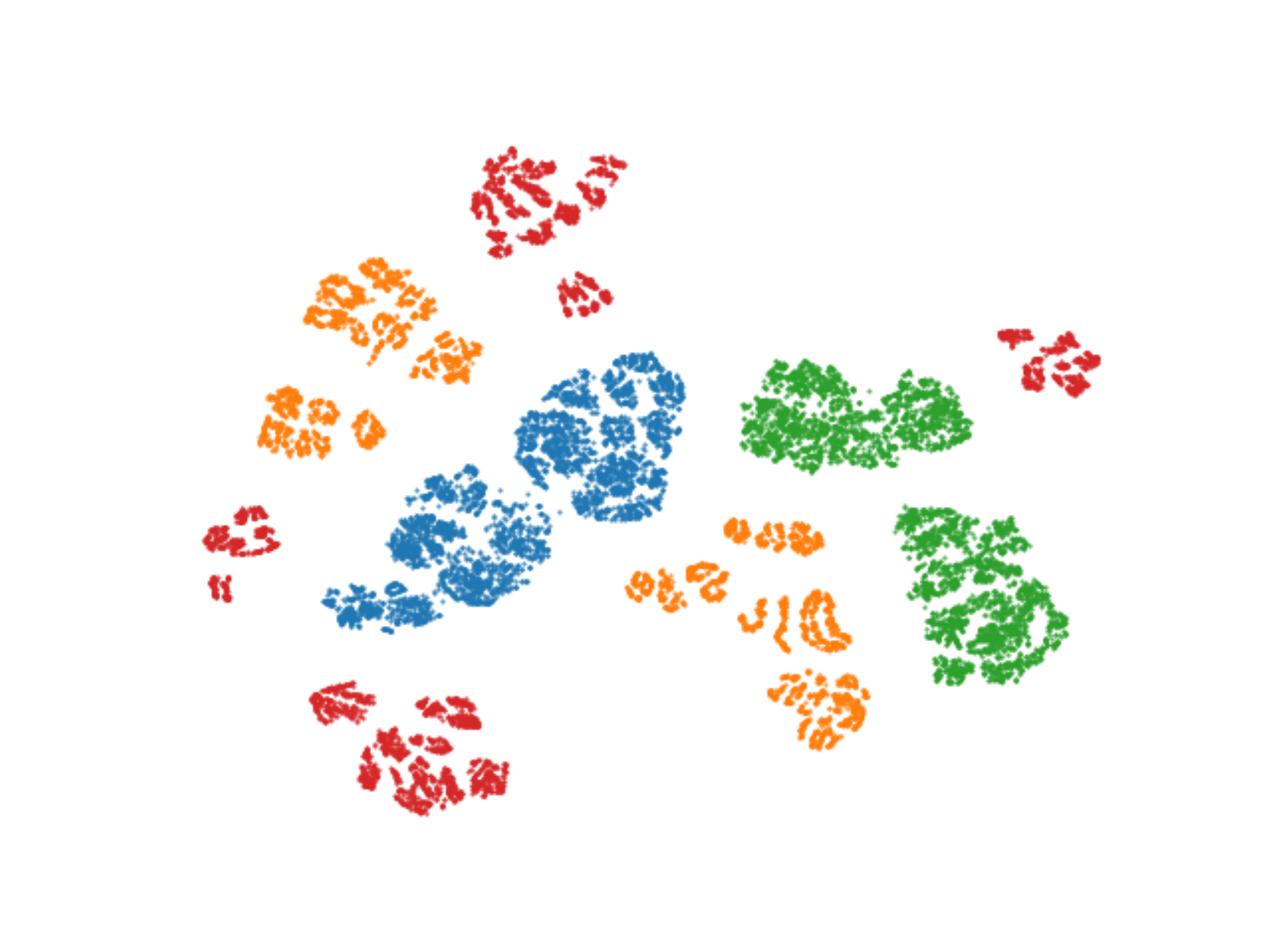}}\\
    \subfloat[Sprites]{
        \includegraphics[width=0.3\linewidth]{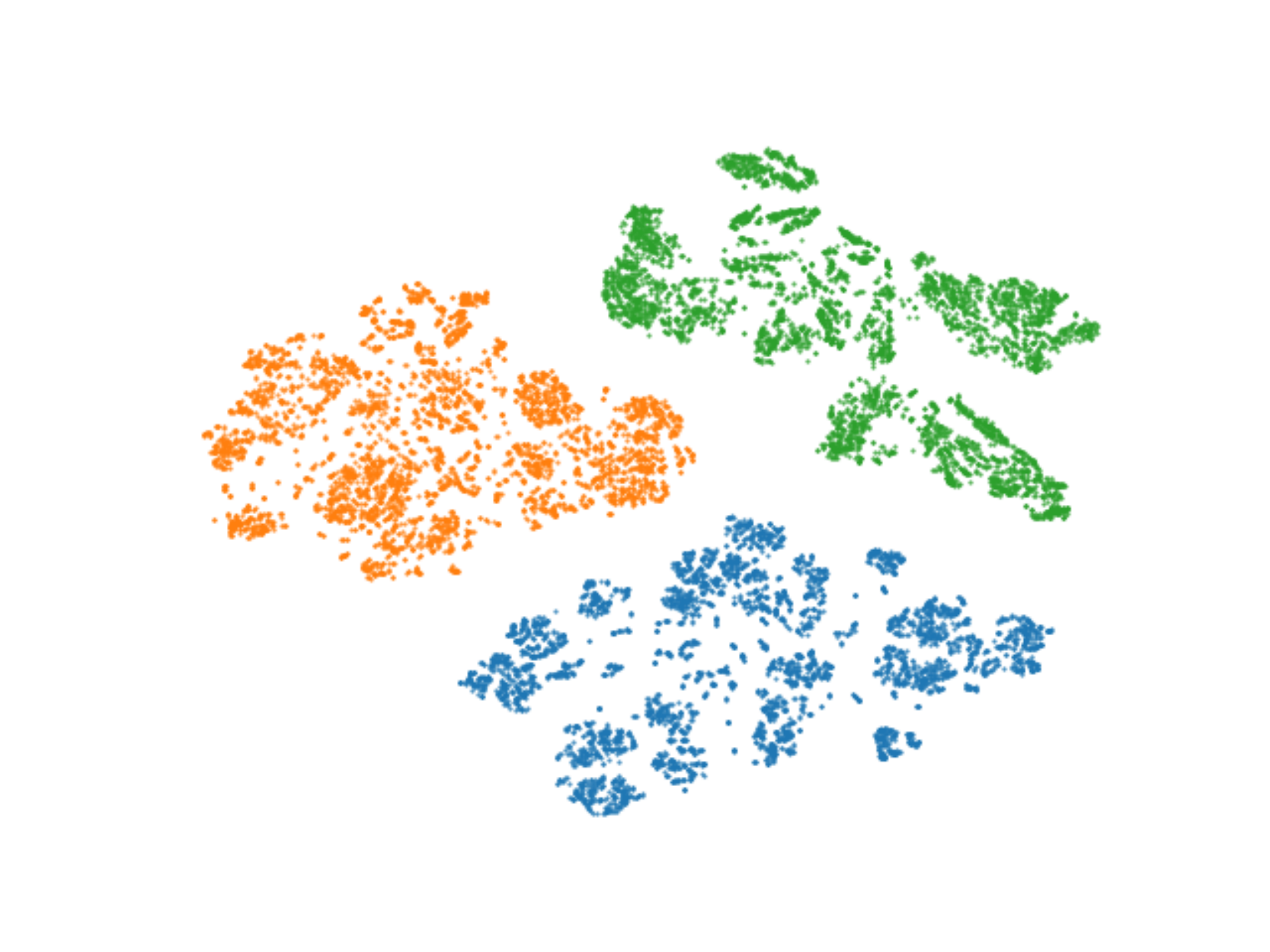}
        \includegraphics[width=0.3\linewidth]{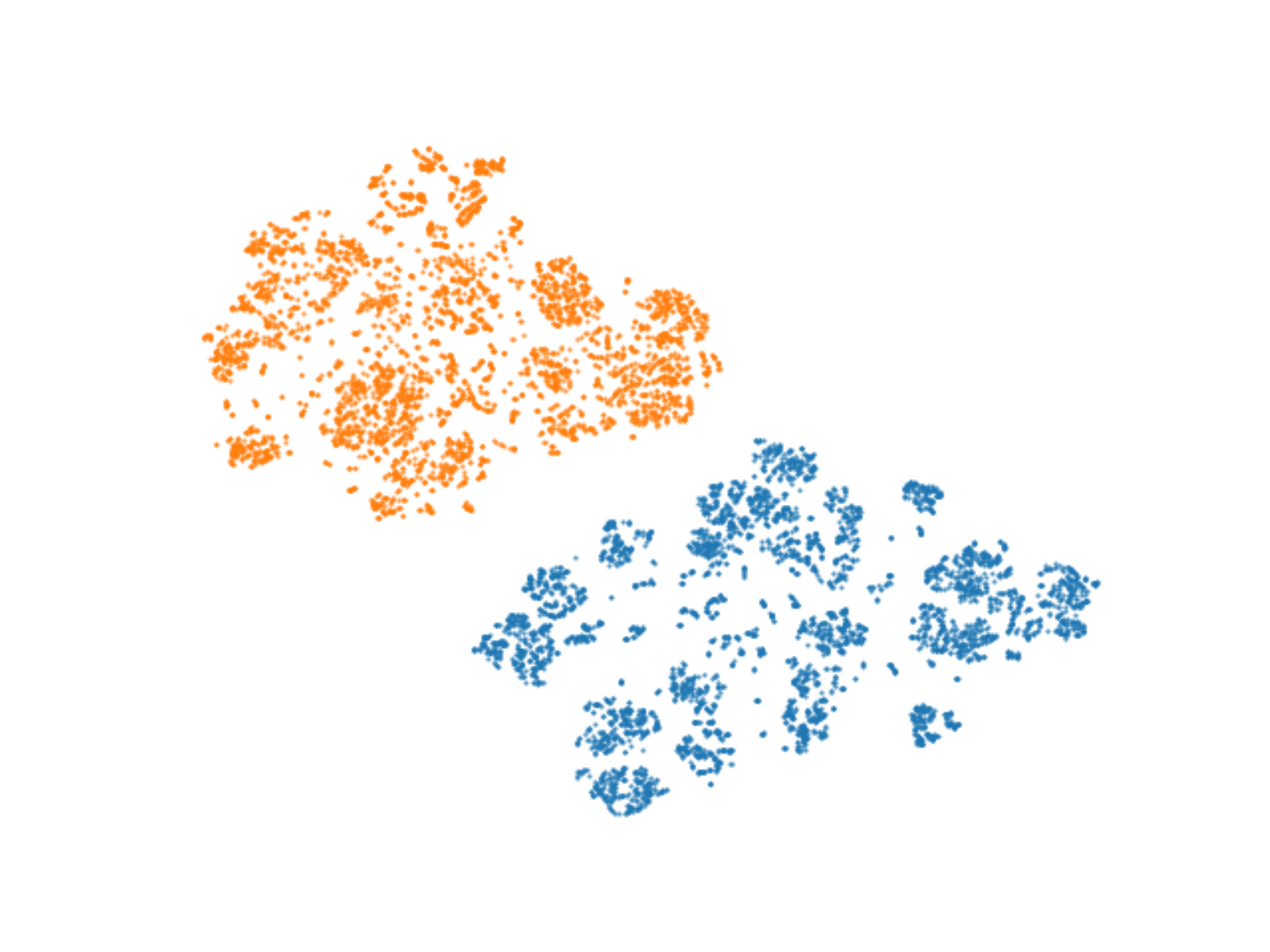}}
    \caption{Latent space visualizations for Cars3D, 3DShapes and Sprites datasets with all feature chunks (Left) \textit{i.e.}, $Z_{f} \cup Z_{u}$ and specified feature chunks (Right) \textit{i.e.}, $Z_{f}$. Each color depicts samples from different latent chunks, green color representing the unspecified chunk, \textit{i.e.}, $Z_{u}$ for the Cars3D and Sprites dataset and purple for the 3DShapes dataset.}
    \label{fig:tsne_appendix}
\end{figure}

In this section, we present the additional latent visualisations of the test set samples with and without the unspecified chunk, \textit{i.e.}, $Z_{u}$. The latent visualizations for the Cars3D, Sprites and 3DShapes datasets are shown in Fig \ref{fig:tsne_appendix}.

\section{Future Work}

In the future, we would like to explore research directions that involve generalizing the set of augmentations used. Further as claimed in \cite{loca}, we would like to evaluate the performance gains of leveraging our disentanglement model in terms of sample complexity on various downstream tasks.

\end{document}